\documentclass[runningheads]{llncs}

\usepackage{eccv}

\usepackage{eccvabbrv}

\usepackage{graphicx}
\usepackage{booktabs}
\usepackage{multirow}
\usepackage{algorithm}
\usepackage{algpseudocode}
\usepackage{todonotes}      
\usepackage{adjustbox}      
\usepackage{caption}
\usepackage{pifont}
\usetikzlibrary{arrows.meta, calc, decorations.pathreplacing}
\usetikzlibrary{arrows.meta, calc}

\usepackage[accsupp]{axessibility}  

\usepackage{hyperref}

\usepackage{orcidlink}

\newcommand{\dt}{\Delta t}

\newcommand{\alon}{a^{\mathrm{lon}}}
\newcommand{\alat}{a^{\mathrm{lat}}}
\newcommand{\Alon}{\mathcal{A}^{\mathrm{lon}}}
\newcommand{\Alat}{\mathcal{A}^{\mathrm{lat}}}
\newcommand{\feas}{\mathcal{F}}
\newcommand{\clip}{\operatorname{clip}}
\newcommand{\ddelta}{\dot{\delta}} 
\newcommand{\jlon}{j^{\mathrm{lon}}}
\newcommand{\jlat}{j^{\mathrm{lat}}}

\definecolor{FeasBlue}{HTML}{0072B2}
\definecolor{AnchorOrange}{HTML}{E69F00}
\definecolor{ClipRed}{HTML}{D55E00}

\newcommand{\toolname}{\textsc{PufferDrive-Editor}}

\newif\ifshowlink
\showlinkfalse 

\begin{document}

\title{Comfort by Construction: Adaptive, Comfort-Bounded Action Spaces for Learned Driving Policies}

\titlerunning{Adaptive, Comfort-Bounded Action Spaces for Learned Driving}

\author{Anna Rothenhäusler\inst{1}\thanks{These authors contributed equally to this work.}\orcidlink{0009-0007-5613-842X} \and
Daniel Jost\inst{1}$^{\star}$\orcidlink{0009-0002-0015-4184} \and
Raghu Rajan\inst{1}\orcidlink{0000-0003-3203-8401} \and
Faris Janjos\inst{2} \and
Oliver Scheel\inst{2} \and
Andreas Look\inst{3}\orcidlink{0000-0002-7423-4091} \and
Joschka Boedecker\inst{1}\orcidlink{0000-0002-3486-7345}}

\authorrunning{A. Rothenhäusler et al.}

\institute{University of Freiburg, Freiburg, Germany\\
\email{\{rothenha, jostd, rajanr, jboedeck\}@cs.uni-freiburg.de} \and
Bosch Center for Artificial Intelligence, Stuttgart, Germany\\
\email{\{faris.janjos, oliver.scheel\}@de.bosch.com} \and
Coburg University of Applied Sciences, Coburg, Germany\\
\email{andreas.look@hs-coburg.de}}


\maketitle


\begin{abstract}
Data-driven driving simulators command accelerations and steering rates from a fixed
grid \emph{without} constraining the \emph{realized} accelerations and jerks. As a
result, reinforcement-learning policies inflate safety metrics through abrupt,
last-second maneuvers that lie far outside the range of human driving and would be
unacceptable to occupants of a real vehicle, so the metrics measure simulator
permissiveness rather than policy quality. Enforcing comfort bounds
naively is not enough: lateral limits shrink quadratically with speed, so clamping a
static grid saturates it and destroys fine-grained control (``grid collapse''). We
propose an \emph{adaptive} action parameterization that re-discretizes the grid at every
step to span exactly the per-step feasible control set, via closed-form inversion of the
lateral-jerk constraint. We further present \toolname, a browser-based tool to audit
realized kinematics and author kinematically challenging scenes. On the Waymo Open Motion
Dataset and a hand-authored slalom, our adaptive model holds comfort violations below
$1\%$ while outperforming clipped-grid and direct-jerk baselines in navigability.
\end{abstract}

\section{Introduction}
\label{sec:intro}
Learned driving policies are increasingly trained in data-driven simulators like Nocturne~\cite{vinitsky2022nocturne}, Waymax~\cite{gulino2023waymax}, or GPUDrive~\cite{kazemkhani2024gpudrive}. These simulators are built for closed-loop reinforcement learning (RL): RL trains on policy-induced states, enabling learning from rare safety-critical situations. While achieving high goal completion, their behavior is shaped by the kinematic bicycle model~\cite{kong2015kinematic,polack2017kinematic} translating actions into motion. Standard interfaces use a fixed grid of accelerations and steering rates that does not constrain the \emph{realized} accelerations and jerks. Consequently, RL policies learn to avoid collisions through abrupt, last-second maneuvers (sharp braking combined with hard steering) whose realized accelerations and, especially, jerks fall well outside the comfort envelope of even aggressive human driving~\cite{bae2020comfort,feng2017jerk}. Safety metrics thus measure the permissiveness of the simulator's kinematics rather than policy quality, crediting evasions that no comfort-respecting vehicle would reproduce. This work studies the problem in \textsc{PufferDrive}, whose actuation interface is representative of the simulators above; the analysis and the proposed parameterization carry over to any simulator built on the same kinematic bicycle model.
Naively enforcing comfort bounds is difficult because the feasible action set depends nonlinearly on the vehicle state. In particular, the admissible steering range shrinks rapidly with speed, alongside additional nonlinearities from the vehicle kinematics~\cite{kong2015kinematic,bae2020comfort}. A static grid tuned for low speed then falls largely outside this range at higher speeds, and clamping collapses many grid points onto the constraint boundary, degrading fine-grained control (``grid collapse''). Because widely used data-driven simulators expose \emph{discrete} action grids~\cite{gulino2023waymax,kazemkhani2024gpudrive}, we focus on maintaining a feasible discrete grid without collapse. To inspect these failure modes, we present \toolname, a browser-based tool for visualizing kinematic violations and authoring challenging scenarios. It lets us trace where policies breach the comfort envelope: qualitatively, most often just before sharp turns and narrow gaps. To keep control feasible in exactly these situations without collapsing the grid, we propose an adaptive action parameterization that computes the exact feasible control set at each step via closed-form inversion of the lateral-jerk constraint, re-discretizing the grid over it.

Our reference points are the established kinematics models used in \textsc{PufferDrive}: \textsc{Classic}, which commands acceleration and steering from a fixed grid and enforces no comfort limits at all, and \textsc{Jerk}, which commands jerk and integrates it into bounded accelerations. We take the steering \emph{rate} as the default steering command, as every constrained model below uses it, and report the original steering-angle parameterization as \textsc{Classic} (steer angle) alongside it. Against them we put three improved variants: \textsc{Jerk} equipped with realistic comfort bounds, and two versions of \textsc{Classic} that resolve their commands against the per-step feasible set: one by \emph{clipping} the fixed grid into it (\textsc{Clipped}), one by \emph{re-discretizing} the grid over it (\textsc{Adaptive}).

Our contributions are as follows:
\begin{enumerate}
\item We show that unconstrained action spaces let learned policies inflate safety metrics through maneuvers that leave the comfort envelope of even aggressive human driving, and that naive clamping leads to grid collapse.
\item We propose a constraint-aware action parameterization that adaptively re-discretizes the per-step feasible control set.
\item We present \toolname, a browser-based tool for auditing realized kinematics and authoring targeted scenarios, providing a general debugging tool for exposing and analyzing failure cases in \textsc{PufferDrive}.
\end{enumerate}

\section{Problem Description}
\label{sec:background_related}

We first describe the vehicle model that turns actions into motion
(Sec.~\ref{sec:bicycle}), then the behavioral comfort envelope that a
realized trajectory should respect (Sec.~\ref{sec:bg-comfort}), before
reviewing how prior work enforces such constraints (Sec.~\ref{sec:related})
and showing why the two action spaces PufferDrive contains fail to do so
(Sec.~\ref{sec:baselines}).

\subsection{Kinematic bicycle model}
\label{sec:bicycle}

\begin{figure}[t]
  \centering
  \begin{tikzpicture}[>=stealth, scale=1.25]
    \draw[->, black!60] (-0.8,0.2) -- (0.2,0.2) node[below] {\footnotesize $x$};
    \draw[->, black!60] (-0.8,0.2) -- (-0.8,1.2) node[left] {\footnotesize $y$};
    \coordinate (R) at (1.6,0.5);
    \coordinate (F) at ($(R)+(22:3.4)$);
    \coordinate (M) at ($(R)!0.5!(F)$);

    \draw[line width=0.8pt] (R) -- (F);

    \draw[line width=2.8pt, black!75] ($(R)+(202:0.42)$) -- ($(R)+(22:0.42)$);
    \draw[line width=2.8pt, black!75] ($(F)+(234:0.42)$) -- ($(F)+(54:0.42)$);

    \draw[dashed, black!50] (R) -- +(1.6,0);
    \draw[dashed, black!50] (F) -- +(22:1.2);

    \draw[black!70] ($(R)+(1.05,0)$)
      arc[start angle=0, end angle=22, radius=1.05];
    \node at ($(R)+(11:1.55)$) {\footnotesize $\psi_t$};

    \draw[black!70] ($(F)+(22:0.8)$)
      arc[start angle=22, end angle=54, radius=0.8];
    \node at ($(F)+(38:1.05)$) {\footnotesize $\delta_t$};

    \draw[dashed, black!50] (F) -- +(54:1.25);

    \draw[->, AnchorOrange, thick] (R) -- +(22:1.8)
      node[above right=4pt] {\footnotesize $v_t$};

    \fill[AnchorOrange] (R) circle (1.5pt);
    \node[below left=4pt] at ($(R)+(-0.05,-0.08)$)
      {\footnotesize $(x_t,y_t)$};

    \draw[<->, black!70] 
      ($(R)+(292:0.45)$) -- ($(F)+(292:0.45)$);
    
    \node[black!70, rotate=22] at ($(M)+(292:0.75)$)
      {\footnotesize $L$};

  \end{tikzpicture}
  \caption{Rear-axle kinematic bicycle model. Each axle is collapsed into a
  single wheel, separated by the wheelbase $L$. The state is the rear-axle
  position $(x_t,y_t)$ (orange), heading $\psi_t$, signed speed $v_t$, and
  steering angle $\delta_t$. The rear wheel does not slip, so the velocity
  (orange) points along the heading $\psi_t$ and $v_t$ is the rear axle's own
  speed. Comfort is judged at this point.}
  \label{fig:bicycle}
\end{figure}
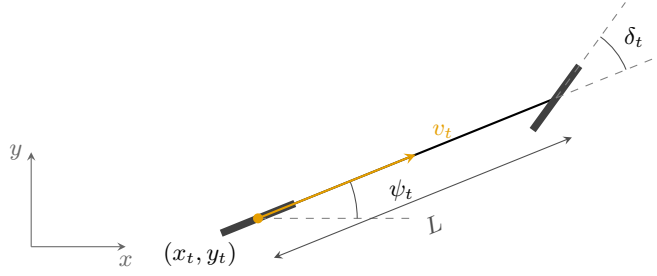

Vehicle motion is commonly abstracted by the \emph{kinematic bicycle
model}~\cite{rajamani2011vehicle, kong2015kinematic, polack2017kinematic}:
the two wheels of each axle are collapsed into a single wheel on the
longitudinal axis, tire slip is neglected, and the vehicle is described by
its pose, speed, and front-wheel steering angle
(Fig.~\ref{fig:bicycle}). Despite its simplicity it tracks real vehicle
trajectories closely in the moderate lateral-acceleration regime of everyday
driving~\cite{polack2017kinematic}, which makes it the standard choice for
planning and for driving simulators.

We parametrize the model at the \emph{rear axle}. This is the reference
point at which the tire-slip-free kinematic bicycle is exact: the rear
wheel cannot slip sideways, so the rear axle's velocity always points along
the vehicle heading and its speed is simply the state speed $v$ --- no slip
angle appears. Choosing it as the reference point makes the comfort
quantities of Sec.~\ref{sec:bg-comfort} identical across every dynamics model
we compare, including those that never model slip at all.

All controlled agents follow this model with state
$(x_t, y_t, \psi_t, v_t, \delta_t)$ as visualized in Fig.~\ref{fig:bicycle}. Given a commanded acceleration $a_t$ and
a steering rate $\ddelta_t$, one
step of length $\dt$ reads
\begin{align}
  v_{t+1} &= \clip\!\big(v_t + a_t\dt,\ \pm v_{\max}\big), &
  \delta_{t+1} &= \clip\!\big(\delta_t + \ddelta_t\dt,\ \pm\delta_{\max}\big), \nonumber\\
  \omega_{t+1} &= \frac{v_t\tan\delta_{t+1}}{L}, &
  \psi_{t+1} &= \psi_t + \omega_{t+1}\dt,
  \label{eq:bicycle}
\end{align}
with yaw rate $\omega$ and wheelbase $L$.

Comfort is judged on the \emph{realized}, not the commanded, longitudinal
and lateral accelerations, and both are read at the rear axle. The
longitudinal acceleration is the change in rear-axle speed, and the lateral
(centripetal) acceleration is that speed times the rear-axle yaw rate:
\begin{equation}
  \alon_{t+1} = \frac{v_{t+1} - v_t}{\dt}, \qquad
  \alat_{t+1} = v_{t+1}\,\omega_{t+1}, \qquad
  \omega_{t+1} = \frac{v_t\tan\delta_{t+1}}{L},
  \label{eq:realized}
\end{equation}
with the realized jerks the backward differences
$\jlon_{t+1} = (\alon_{t+1}-\alon_{t})/\dt$ and
$\jlat_{t+1} = (\alat_{t+1}-\alat_{t})/\dt$.

\subsection{Comfort bounds on vehicle kinematics}
\label{sec:bg-comfort}

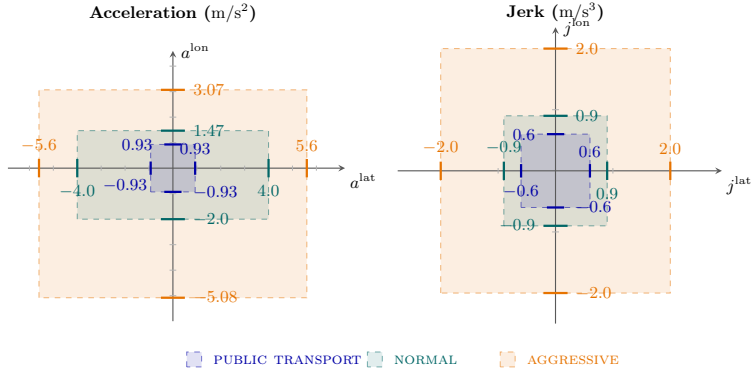
\begin{figure}[t]
  \centering
  \resizebox{0.82\linewidth}{!}{
\begin{tikzpicture}[>=stealth,
    agg/.style={orange!90!black},
    nrm/.style={teal!80!black},
    pub/.style={blue!65!black},
    axisline/.style={->, semithick, black!65},
    minortick/.style={black!30, thin},
    bndlab/.style={font=\footnotesize},
    region/.style={draw, dashed, thin, fill opacity=0.12, draw opacity=0.55}]

  \begin{scope}[x=0.45cm, y=0.48cm]

    \fill[region, agg, fill=orange!90!black] (-5.6,-5.08) rectangle (5.6,3.07);
    \fill[region, nrm, fill=teal!80!black]   (-4.0,-2.00) rectangle (4.0,1.47);
    \fill[region, pub, fill=blue!65!black]   (-0.93,-0.93) rectangle (0.93,0.93);

    \draw[axisline] (-6.9,0) -- (7.2,0)
      node[below right=-1pt, font=\small, black] {$a^{\mathrm{lat}}$};
    \draw[axisline] (0,-6.0) -- (0,4.6)
      node[right=1pt, font=\small, black] {$a^{\mathrm{lon}}$};
    \foreach \t in {-6,-5,-4,-3,-2,-1,1,2,3,4,5,6} {
      \draw[minortick] (\t,-0.10) -- (\t,0.10);
    }
    \foreach \t in {-5,-4,-3,-2,-1,1,2,3,4} {
      \draw[minortick] (-0.14,\t) -- (0.14,\t);
    }

    \foreach \v in {-0.93,0.93} {
      \draw[pub, very thick] (\v,-0.26) -- (\v,0.26);
    }
    \foreach \v in {-4.0,4.0} {
      \draw[nrm, very thick] (\v,-0.36) -- (\v,0.36);
    }
    \foreach \v in {-5.6,5.6} {
      \draw[agg, very thick] (\v,-0.36) -- (\v,0.36);
    }
    \node[bndlab, pub, above] at ( 0.93, 0.30) {$0.93$};
    \node[bndlab, pub, below left=-2pt] at (-0.93,-0.30) {$-0.93$};
    \node[bndlab, nrm, below] at (-4.0, -0.42) {$-4.0$};
    \node[bndlab, nrm, below] at ( 4.0, -0.42) {$4.0$};
    \node[bndlab, agg, above] at (-5.6,  0.42) {$-5.6$};
    \node[bndlab, agg, above] at ( 5.6,  0.42) {$5.6$};

    \foreach \v in {-0.93,0.93} {
      \draw[pub, very thick] (-0.36,\v) -- (0.36,\v);
    }
    \foreach \v in {-2.00,1.47} {
      \draw[nrm, very thick] (-0.5,\v) -- (0.5,\v);
    }
    \foreach \v in {-5.08,3.07} {
      \draw[agg, very thick] (-0.5,\v) -- (0.5,\v);
    }
    \node[bndlab, pub, left]  at (-0.6, 0.93) {$0.93$};
    \node[bndlab, pub, right] at ( 0.6,-0.93) {$-0.93$};
    \node[bndlab, nrm, right] at ( 0.6, 1.47) {$1.47$};
    \node[bndlab, nrm, right] at ( 0.6,-2.00) {$-2.0$};
    \node[bndlab, agg, right] at ( 0.6, 3.07) {$3.07$};
    \node[bndlab, agg, right] at ( 0.6,-5.08) {$-5.08$};

    \node[font=\small\bfseries, black] at (0,6.1)
      {Acceleration ($\mathrm{m/s^2}$)};
  \end{scope}

  \begin{scope}[shift={(7.2cm,-0.05cm)}, x=1.08cm, y=1.15cm]

    \fill[region, agg, fill=orange!90!black] (-2.0,-2.0) rectangle (2.0,2.0);
    \fill[region, nrm, fill=teal!80!black]   (-0.9,-0.9) rectangle (0.9,0.9);
    \fill[region, pub, fill=blue!65!black]   (-0.6,-0.6) rectangle (0.6,0.6);

    \draw[axisline] (-2.75,0) -- (2.9,0)
      node[below right=-1pt, font=\small, black] {$j^{\mathrm{lat}}$};
    \draw[axisline] (0,-2.5) -- (0,2.35)
      node[right=1pt, font=\small, black] {$j^{\mathrm{lon}}$};
    \foreach \t in {-2,-1,1,2} {
      \draw[minortick] (\t,-0.045) -- (\t,0.045);
      \draw[minortick] (-0.06,\t) -- (0.06,\t);
    }

    \foreach \v in {-0.6,0.6} {
      \draw[pub, very thick] (\v,-0.11) -- (\v,0.11);
    }
    \foreach \v in {-0.9,0.9} {
      \draw[nrm, very thick] (\v,-0.15) -- (\v,0.15);
    }
    \foreach \v in {-2.0,2.0} {
      \draw[agg, very thick] (\v,-0.15) -- (\v,0.15);
    }
    \node[bndlab, pub, above] at ( 0.6, 0.13) {$0.6$};
    \node[bndlab, pub, below] at (-0.6,-0.13) {$-0.6$};
    \node[bndlab, nrm, below] at ( 0.9,-0.18) {$0.9$};
    \node[bndlab, nrm, above] at (-0.9, 0.18) {$-0.9$};
    \node[bndlab, agg, above] at ( 2.0, 0.18) {$2.0$};
    \node[bndlab, agg, above] at (-2.0, 0.18) {$-2.0$};

    \foreach \v in {-0.6,0.6} {
      \draw[pub, very thick] (-0.15,\v) -- (0.15,\v);
    }
    \foreach \v in {-0.9,0.9} {
      \draw[nrm, very thick] (-0.21,\v) -- (0.21,\v);
    }
    \foreach \v in {-2.0,2.0} {
      \draw[agg, very thick] (-0.21,\v) -- (0.21,\v);
    }
    \node[bndlab, pub, left]  at (-0.25, 0.6) {$0.6$};
    \node[bndlab, pub, right] at ( 0.25,-0.6) {$-0.6$};
    \node[bndlab, nrm, right] at ( 0.25, 0.9) {$0.9$};
    \node[bndlab, nrm, left]  at (-0.25,-0.9) {$-0.9$};
    \node[bndlab, agg, right] at ( 0.25, 2.0) {$2.0$};
    \node[bndlab, agg, right] at ( 0.25,-2.0) {$-2.0$};

    \node[font=\small\bfseries, black] at (0,2.60)
      {Jerk ($\mathrm{m/s^3}$)};
  \end{scope}

  \begin{scope}[shift={(0.4cm,-3.6cm)}]
    \fill[region, pub, fill=blue!65!black] (-0.14,-0.14) rectangle (0.14,0.14);
    \node[right, font=\footnotesize, pub] at (0.25,0) {\textsc{public transport}};
    \fill[region, nrm, fill=teal!80!black] (3.26,-0.14) rectangle (3.54,0.14);
    \node[right, font=\footnotesize, nrm] at (3.65,0) {\textsc{normal}};
    \fill[region, agg, fill=orange!90!black] (5.76,-0.14) rectangle (6.04,0.14);
    \node[right, font=\footnotesize, agg] at (6.15,0) {\textsc{aggressive}};
  \end{scope}

\end{tikzpicture}}
  \caption{Comfort limits of the OPM driver profiles of~\cite{bae2020comfort}, marked on each axis; each profile lies strictly inside the next. We use \textsc{normal} and \textsc{aggressive} in our experiments and show \textsc{public transport} for reference. (Left)~Acceleration plane $(\alat,\alon)$ in $\mathrm{m/s^2}$; the longitudinal limits of the driving profiles are asymmetric, since braking tolerance exceeds throttle tolerance. (Right)~Jerk plane $(\jlat,\jlon)$ in $\mathrm{m/s^3}$, with symmetric limits.}
  \label{fig:comfort_bounds}
\end{figure}
Occupants perceive vehicle motion through acceleration and its time derivative,
jerk; abrupt or sustained accelerations can cause discomfort and motion
sickness~\cite{dewinkel2023comfort,elbanhawi2015comfort}. We adopt the
Occupant's Preference Metric (OPM) of Bae~\etal~\cite{bae2020comfort}, which
defines a behavioral comfort envelope via five parameters:
\begin{equation}
\mathrm{OPM} \;=\; \big\{\,
a^{\mathrm{lon}}_{\max},\;
a^{\mathrm{lon}}_{\min},\;
a^{\mathrm{lat}}_{\max},\;
j^{\mathrm{lon}}_{\max},\;
j^{\mathrm{lat}}_{\max}
\,\big\},
\label{eq:opm}
\end{equation}
where $a^{\mathrm{lon}}_{\max}$ and $a^{\mathrm{lon}}_{\min}$ denote the
throttle and braking thresholds, respectively, while the lateral acceleration
and jerk limits are symmetric about zero. Accordingly, the five OPM parameters
define lower and upper bounds on the four realized quantities of
Sec.~\ref{sec:bicycle}:
\[
a^{\bullet} \in [\underline{a}^{\bullet}, \bar{a}^{\bullet}],
\qquad
j^{\bullet} \in [\underline{j}^{\bullet}, \bar{j}^{\bullet}],
\qquad
\bullet \in \{\mathrm{lon},\mathrm{lat}\},
\]
with
$\underline{a}^{\mathrm{lon}} = a^{\mathrm{lon}}_{\min}$,
$\bar{a}^{\mathrm{lon}} = a^{\mathrm{lon}}_{\max}$,
$\underline{a}^{\mathrm{lat}} = -a^{\mathrm{lat}}_{\max}$,
$\bar{a}^{\mathrm{lat}} = a^{\mathrm{lat}}_{\max}$, and
$\underline{j}^{\bullet} = -j^{\bullet}_{\max}$,
$\bar{j}^{\bullet} = j^{\bullet}_{\max}$.
Figure~\ref{fig:comfort_bounds} illustrates these limits. Bae~\etal{} anchor the tightest profile in public-transport data, where freestanding passengers lose posture~\cite{martin2008jerk}; we show it for reference but drive with the two automotive profiles. The \textsc{normal} profile uses seated-occupant thresholds, and the \textsc{aggressive} profile uses observed human driving extremes; naturalistic driving studies likewise treat sustained high jerk not as a physical impossibility but as the signature of an aggressive driver~\cite{feng2017jerk}.

These are \emph{behavioral} envelopes, sitting well inside the physical handling limits: leaving the \textsc{aggressive} band is not impossible, only something essentially no human driver would choose and no occupant would accept. We therefore call a control \textsc{feasible} relative to the enforced profile rather than in an absolute physical sense.

In the original OPM system, bounds are soft constraints enforced at the path-planning level~\cite{bae2020comfort}. Data-driven traffic simulators also evaluate these comfort quantities \emph{post hoc} using metrics or Wasserstein distance penalties~\cite{xu2023bits,montali2023wosac}. In contrast, we use the comfort envelope to constrain the policy's action space directly: at each step, we analytically compute the feasible set of controls whose realized body-frame quantities stay inside the envelope (Figure~\ref{fig:comfort_bounds}) and re-discretize the action grid over this set (Section~\ref{sec:feasible}), guaranteeing comfort by construction.

\subsection{Kinematic constraints in simulation and reinforcement learning}
\label{sec:related}

\paragraph{Kinematic models in driving simulators.}
Data-driven simulators such as Nocturne~\cite{vinitsky2022nocturne}, Waymax~\cite{gulino2023waymax}, and GPUDrive~\cite{kazemkhani2024gpudrive} share a common actuation interface: a kinematic bicycle model~\cite{kong2015kinematic,polack2017kinematic} driven from a fixed grid of controls that is clamped only to static input ranges. Nothing bounds the realized accelerations or jerks, the gap our work targets in Section \ref{sec:methods}.

\paragraph{Feasibility in learned models.}
Prior works build kinematic constraints into learned motion models, such as routing trajectories through a bicycle-model layer (e.g., Deep Kinematic Models~\cite{cui2020deepkinematic}, Trajectron++~\cite{salzmann2020trajectronpp}) or propagating uncertainty through stochastic kinematic layers~\cite{westny2024stochastic}. While these methods make \emph{predicted trajectories} dynamically consistent by construction, they act downstream of action selection: a policy remains free to command an out-of-envelope maneuver that the layer then realizes rather than forbids. We instead constrain selection itself, bounding the feasible action set of a \emph{closed-loop policy} at every simulation step under a behavioral comfort envelope. We expect this to matter most for the adaptive variant: by re-discretizing a full-resolution grid over the feasible set, it lets the policy optimize entirely within the envelope, so the constraint is reflected in what the policy learns rather than corrected after the fact.

\paragraph{Constrained reinforcement learning.}
The simplest approach leaves the action space untouched and penalizes violations in the reward, as Gigaflow~\cite{cusumano2025selfplay} does for harsh acceleration and jerk. Comfort then competes with progress through a tunable weight: violations get rarer but are never ruled out. Hard enforcement is typically done via external layers that mask or project actions~\cite{dalal2018safe,cheng2019cbf,krasowski2020safe,wang2021commonroadrl}. However, when the feasible set shrinks (e.g., quadratically with speed), these methods suffer from grid collapse, where distinct discrete actions alias to the same boundary values. While action mapping~\cite{theile2025actionmapping} and jerk-bounded generators~\cite{lunardi2024jerkbounded} address constrained action selection, our approach analytically computes a per-step feasible action region and adaptively re-discretizes the action grid onto it, preserving action resolution across varying vehicle speeds while guaranteeing feasible commands.

\subsection{Default models and their failure modes}
\label{sec:baselines}

PufferDrive provides two kinematics models by default. \textsc{Classic}
commands an acceleration and a steering command directly from a fixed
$7\times13$ grid, applied unchanged. Nothing ties consecutive commands
together: within one step of $\dt = 0.1\,\mathrm{s}$ the policy may switch
from full braking to full throttle or slew the steering across its whole
range, producing jerks orders of magnitude beyond any human comfort
envelope. As a result, \textsc{Classic}
learns effective but kinematically implausible driving.

The steering command comes in two parameterizations: the steering
\emph{angle} $\delta_t$, as originally implemented, or the steering
\emph{rate} $\ddelta_t$, which integrates to the angle and therefore
low-pass filters the steering without bounding anything. Every constrained
model in this work commands the rate (Sec.~\ref{sec:discussion}), so we treat
it as the default parameterization: \textsc{Classic} denotes the steer-rate
model throughout, and the original is written \textsc{Classic} (steer angle).

\textsc{Jerk} sits at the opposite end: the policy commands longitudinal and
lateral \emph{jerk}, which the environment integrates into accelerations
clamped to the comfort band, so the limits hold by rule. It is not a separate
dynamics model: the same kinematic bicycle model of Sec.~\ref{sec:bicycle}
runs underneath, only re-parameterized one derivative up. The longitudinal
jerk integrates to the speed, while the lateral jerk integrates to a lateral
acceleration that is mapped back to a steering angle through the bicycle
curvature relation $\delta = \arctan(\kappa L)$ with $\kappa = \alat/v^2$. The
disadvantage is that the action now operates on the second derivative of the
velocity:
credit assignment must propagate through $j \to a \to v \to x$, and the same
jerk action produces different motion depending on the accumulated
acceleration state. A single action barely changes the trajectory, and
exploration in jerk space is much harder.

\section{Method}
\label{sec:methods}

We therefore introduce an action space in between: the policy commands an
\emph{acceleration and a steering rate} $(a_t, \ddelta_t)$ --- one derivative closer
to the trajectory than jerk control --- and the comfort limits are enforced
inside the environment step by restricting each command to the set feasible
from the current state (Sec.~\ref{sec:feasible}). Two variants of this
restriction are compared (Sec.~\ref{sec:variants}): \emph{clipping}, which
keeps \textsc{Classic}'s absolute action semantics, and \emph{adaptive
re-discretization}, which re-maps the action grid onto the feasible set at each
step so that actions never collapse onto the limit values.

\subsection{Per-step feasible set}
\label{sec:feasible}

Combining the jerk and absolute limits yields per-axis target bands $\Alon_t$ and $\Alat_t$. The longitudinal band directly bounds the commanded acceleration $a_t$ via $v_{t+1}=v_t+a_t\dt$. In contrast, the lateral band couples both control inputs:
\[
\alat_{t+1}(a_t,\ddelta_t)=v_{t+1}(a_t)\,\omega_{t+1}(\ddelta_t),
\]
where
\begin{equation}
\omega_{t+1}(\ddelta_t)=\frac{v_t\tan(\delta_t+\ddelta_t\dt)}{L}.
\label{eq:coupling}
\end{equation}
Thus, the realized lateral acceleration depends on both the next speed, determined by $a$, and the next steering angle, determined by $\ddelta$. The jointly feasible action set is
\begin{equation}
\feas_t=\big\{(a_t,\ddelta_t):\alon_{t+1}(a_t)\in\Alon_{t+1},\ \alat_{t+1}(a_t,\ddelta_t)\in\Alat_{t+1}\big\},
\label{eq:feasible}
\end{equation}
and is generally not separable into independent intervals.

A key observation is that $\omega_{t+1}$ depends only on the steering-rate command and the known current speed $v_t$. Consequently, for fixed $\ddelta_t$, $\alat_{t+1}$ is linear in $v_{t+1}$, allowing the lateral band $\Alat_{t+1}$ to be inverted analytically into an interval on $v_{t+1}$.

\paragraph{Constructing the feasible box.}
The box is a deterministic function of the current state. We first intersect
the actuator-rate limits with the
mechanical angle constraint $|\delta_t+\ddelta_t\dt|\leq\delta_{\max}$.
Sampling this entire interval uniformly is ineffective at speed: the lateral
band permits only a narrow interval of rates. We therefore invert $\Alat_{t+1}$
at the nominal next speed $v_{\mathrm{nom}}=v_t+\alon_t\dt$,
which yields candidate rates
\[
 \ddelta_t = \frac{\arctan\!\left(L\,\alat_{t+1}/(v_t v_{\mathrm{nom}})\right)-\delta_t}{\dt},
 \qquad \alat_{t+1}\in\Alat_{t+1}.
\]
The nominal-speed approximation only guides sampling: we widen the resulting
rate window, intersect it with the hard rate and angle limits, and retain the
full hard interval at low speed. We sample $17$ rates centered on the prior
rate and test each against the exact reachable acceleration interval.

For each sampled rate, $\omega_{t+1}$ is known. Intersecting the speed interval
reachable through the longitudinal band with the interval obtained by
dividing $\Alat_{t+1}$ by $\omega_{t+1}$ gives the exact feasible acceleration
interval for that rate: a vertical slice of $\feas_t$ (Fig.~\ref{fig:box}).
For every contiguous feasible run, we intersect its acceleration intervals to
form a rectangle.
The retained box
\(
[\underline a_t,\bar a_t]\times
[\underline{\ddelta}_t,\overline{\ddelta}_t]
\)
is the candidate with the largest normalized area; when the anchor sample is
feasible, candidates must contain it. Its middle discrete action therefore
repeats the last action whenever compatible with comfort. The box is an
inner approximation of $\feas_t$, recomputed at every control step, and every
command it contains is feasible by construction.

\subsection{Clipped and adaptive action mapping}
\label{sec:variants}

\begin{figure}[t]
  \centering
  \begin{subfigure}[b]{0.48\linewidth}
    \centering
    \begin{adjustbox}{max width=\linewidth}
      \begin{tikzpicture}[>=stealth]
        \input{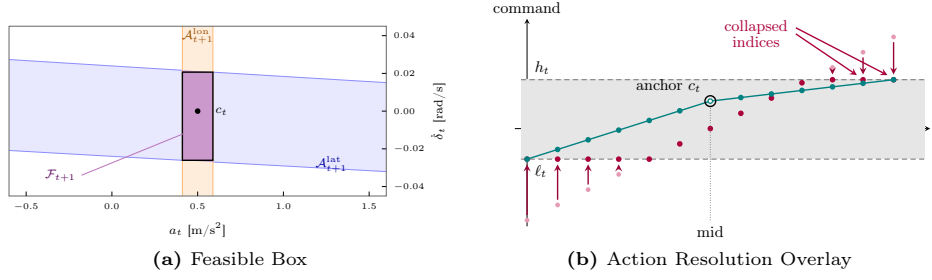}
      \end{tikzpicture}
    \end{adjustbox}
    \caption{Feasible Box}
    \label{fig:box}
  \end{subfigure}\hfill
  \begin{subfigure}[b]{0.48\linewidth}
    \centering
    \begin{adjustbox}{max width=\linewidth}
      \begin{tikzpicture}[scale=0.62, >=stealth]
        \draw[->] (-0.4,0) -- (13.2,0) node[below left=1pt] {\footnotesize index $k$};
        \draw[->] (0,-3.2) -- (0,3.6) node[above] {\footnotesize command};
        
        \fill[black!10] (-0.2,-1.0) rectangle (13.0,1.6);
        \draw[densely dashed, black!55] (-0.2,1.6) -- (13.0,1.6) node[pos=0.02, above right=0pt, black] {\footnotesize $h_t$};
        \draw[densely dashed, black!55] (-0.2,-1.0) -- (13.0,-1.0) node[pos=0.02, below right=0pt, black] {\footnotesize $\ell_t$};
        
        \foreach \k in {0,...,12} {
          \pgfmathsetmacro{\raw}{-3 + 0.5*\k}
          \pgfmathsetmacro{\clipped}{max(min(\raw,1.6),-1.0)}
          \fill[purple!40] (\k,\raw) circle (2.2pt);
          \fill[purple!90!black] (\k,\clipped) circle (2.6pt);
          \pgfmathparse{abs(\raw-\clipped) > 0.01 ? 1 : 0}
          \ifnum\pgfmathresult>0
            \draw[->, thick, purple!80!black, shorten >=3pt, shorten <=3pt] (\k,\raw) -- (\k,\clipped);
          \fi
        }
        \node[align=center, purple!90!black] at (7.5,3.1) {\footnotesize collapsed\\[-2pt]\footnotesize indices};
        \draw[->, thick, purple!90!black] (8.8,3.1) -- (10.8,1.75);
        \draw[->, thick, purple!90!black] (8.8,3.3) -- (11.8,1.75);
        
        \draw[black!60, densely dotted] (6,-3.0) -- (6,0.9);
        \node[below] at (6,-3.0) {\footnotesize mid};
        \draw[teal, thick] (0,-1.0) -- (6,0.9) -- (12,1.6);
        \foreach \k in {0,...,6} {
          \pgfmathsetmacro{\y}{-1.0 + (0.9-(-1.0))*\k/6}
          \fill[teal] (\k,\y) circle (2.6pt);
        }
        \foreach \k in {7,...,12} {
          \pgfmathsetmacro{\y}{0.9 + (1.6-0.9)*(\k-6)/6}
          \fill[teal] (\k,\y) circle (2.6pt);
        }
        \fill[white] (6,0.9) circle (1.5pt);
        \draw[black, thick] (6,0.9) circle (5pt);
        \node[above left=3pt] at (6,0.9) {\footnotesize anchor $c_t$};
      \end{tikzpicture}
    \end{adjustbox}
    \caption{Action Resolution Overlay}
    \label{fig:mapping_overlay}
  \end{subfigure}
  
  \caption{Construction and utilization of the feasible action space. \textbf{(a)} Per-step construction of the feasible action box (computed from an actual vehicle state; see text for parameters). The orange band denotes the longitudinal target interval $\Alon_{t+1}$, while the blue region contains all $(a_t,\ddelta_t)$ pairs satisfying the lateral target band $\Alat_{t+1}$. Their intersection (purple) is the feasible set $\feas_{t+1}$, from which the largest axis-aligned rectangle (black) is retained. \textbf{(b)} Comparison of action resolution strategies against the feasible interval $[\ell_t, h_t]$. In the \textcolor{purple!90!black}{Clipped} approach (purple arrows and dark purple dots), fixed grid values saturate at the borders, causing multiple indices to collapse. In the \textcolor{teal}{Adaptive} approach (teal line and dots), the grid is directly re-discretized onto the interval, locking the middle index to the anchor $c_t$ (circled) and spreading extremes exactly to the borders so that all indices stay distinct and feasible.}
  \label{fig:action_space}
\end{figure}

Both variants draw from the same nominal $7\times13$ grid of accelerations
and steering rates; they differ in what an action index \emph{means}
(Fig.~\ref{fig:mapping_overlay}).

\paragraph{Clipped.}
Action indices keep fixed physical values, and the requested command is
clipped componentwise into the box. Semantics are stable, but whenever the box is smaller than
the grid span --- the typical regime for steering at speed --- many indices
collapse onto the same box border: effectively only the extreme commands
remain distinguishable, and the policy cannot tell from the action alone
what will actually be executed.

\paragraph{Adaptive.}
To ensure a fair comparison with the \textsc{Classic} model in \textsc{PufferDrive}, we use the same number of discrete actions ($7\times13$), while remapping them to acceleration and steering-rate commands. These discrete actions are mapped to acceleration and steering-rate commands and re-discretized onto the current feasible box at every step using a piecewise-linear \emph{anchored map}: for an interval $[\ell_t, h_t]$ and anchor $c_t$, the middle index maps exactly to $c_t$, the extreme indices to the borders, and intermediate indices interpolate linearly on each side. The anchor is chosen as the previous step's \emph{realized} action whenever it remains feasible under the current box; otherwise, the interval is mapped without an anchor. When used, the anchor gives the middle action the semantics ``repeat what you just did''. All $7\times13$ actions remain distinct and feasible by construction --- nothing saturates --- at the price of \emph{relative} semantics: the same index denotes different physical commands in different states.

\subsection{Observations}
\label{sec:obs}

The feasible box depends on the previous realized accelerations and steering
rate, which are not visible in the instantaneous pose --- a policy without
this information would face a partially observed problem, since identical
observations could resolve the same action differently. Both variants
therefore observe, in addition to the standard ego features, the previous
realized action (the anchor) and the four box edges, normalized by the
profile limits. The two variants share the identical observation layout, so
their comparison isolates the action-mapping strategy alone.

\section{Experiments}
\label{sec:experiments}

We evaluate our comfort-bounded kinematic models against unconstrained and naively bounded baselines. Our experiments are designed to address two key questions: \textbf{(1)~Limit Compliance:} does enforcing the comfort envelope actually eliminate envelope-violating motion? \textbf{(2)~Navigation Cost:} what, if any, is the cost of these comfort constraints on navigation performance?

\subsection{Experimental Setup}
We evaluate our models in \textsc{PufferDrive}, a data-driven 2D bird's-eye view simulator built on the Waymo Open Motion Dataset (WOMD)~\cite{ettinger2021womd} and parallelized via PufferLib~\cite{suarez2024pufferlib}. All vehicles in a scene are controlled simultaneously for episodes of $9.1\,$s ($\dt = 0.1\,$s), with agents removed upon collision, leaving the drivable area, or reaching their goal. We train on $80{,}000$ WOMD scenarios and evaluate on the held-out validation split of $10{,}000$ scenarios.

Policies are trained using Proximal Policy Optimization (PPO)~\cite{schulman2017proximal} for $2$ billion steps across three seeds, using a recurrent actor-critic network and simple reward terms such as collision, offroad and goal reaching. All configurations share identical hyperparameters and training budgets, differing only in their action spaces and bound enforcements.

\subsection{Model Configurations}
We compare the six kinematic models introduced above. The three \emph{unconstrained}
baselines, \textsc{Classic}, \textsc{Classic} (steer angle) and \textsc{Jerk}
(Section~\ref{sec:baselines}), check no
action against the comfort envelope (Figure~\ref{fig:comfort_bounds}). The first two differ only in whether the
policy commands the steering rate or the steering angle. The pair isolates the effect of that interface choice from
the effect of enforcing the envelope. The other three enforce the envelope, in
increasing order of how much of the feasible set they preserve: \textsc{Jerk} (bounded
grid) restricts the jerk grid a priori with a single static conservative grid; Clipped
\textsc{Classic} clips each command into the per-step feasible box axis by axis; and
Adaptive \textsc{Classic} (ours) re-discretizes its grid onto that box, so every
selectable action is feasible by construction (Section~\ref{sec:variants}). Every
\textsc{Classic} model except the one qualified as (steer angle) commands steering
\emph{rate} rather than angle (Section~\ref{sec:discussion}).

The comfort envelope is a property of the simulator, not the model: each constrained
model builds its feasible set from either the \textsc{aggressive} or the tighter
\textsc{normal} band (Figure~\ref{fig:comfort_bounds}). Unless stated otherwise, results
enforce the \textsc{aggressive} band.

\subsection{Evaluation Protocol}
\label{sec:eval-protocol}
Each trained policy is evaluated on the validation split, and we report mean $\pm$ standard deviation over the three seeds.

We report three navigation metrics (the share of agents reaching their goal, the share colliding, and the share leaving the drivable area) and, for each of the four kinematic quantities ($\alon, \alat, \jlon, \jlat$), the share of steps that violate the comfort envelope. Two properties of this tally matter for reading the results. First, we measure the \emph{realized} quantity, not the commanded one: a model that bounds its inputs but still drives outside the envelope through the coupling terms is counted as violating, which is precisely the failure mode naive clamping exhibits. Second, the tally covers only the steps an agent actually drives (steps after removal or goal completion are excluded), so the rate is a share of driving rather than of episode length, which would otherwise reward a model for finishing early.

\subsection{Qualitative Analysis and \toolname{}}
\label{sec:qualitative-setup}

\begin{figure}[!t]
    \centering
    \includegraphics[width=0.75\linewidth]{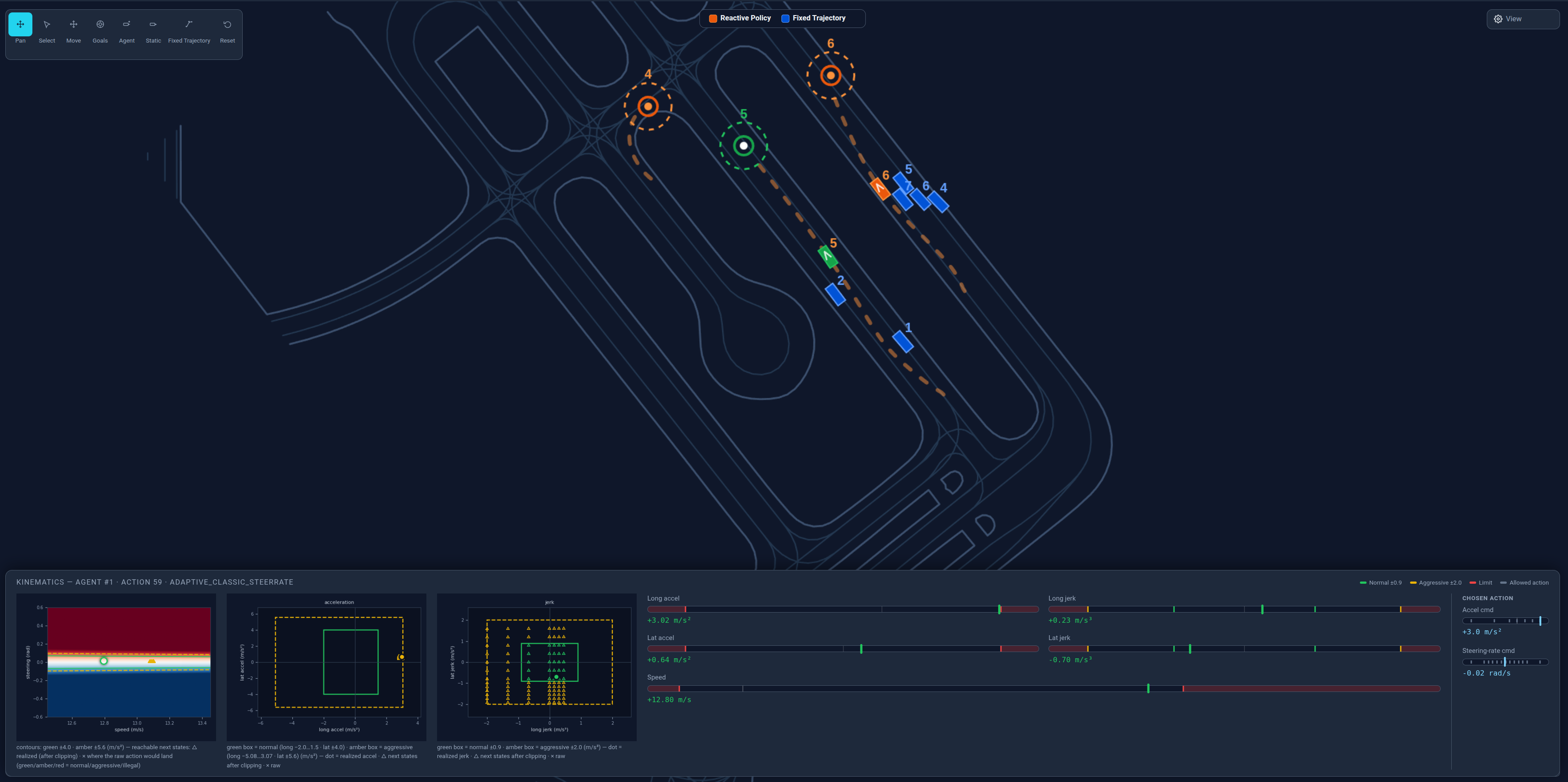}
    \caption{The \toolname{} user interface. The upper panel displays the map canvas and vehicle trajectory paths (e.g., analyzing the reactive policy in orange). The lower panel contains the real-time kinematics dashboard, showing the active passenger-comfort limits (green/amber boxes) for acceleration and jerk, the vehicle's actual realized state trajectory (highlighted dots and trail), and target action re-discretization intervals.}
    \label{fig:editor-screenshot}
\end{figure}

Aggregate metrics over the validation split average away the rare, safety-critical
moments in which the limits actually bind. To inspect behavior near those limits we
present \toolname{}, a browser-based scene editor and policy-debugging interface that
runs locally on the map and scenario binaries loaded by PufferDrive. Although motivated by the analysis of our constraint-aware kinematic models, \toolname{} is a general tool for understanding learned driving policies in PufferDrive, enabling researchers to identify policy strengths, weaknesses, and failure modes beyond aggregate metrics. It combines three functions. A \emph{scenario and track editor} loads road
layouts from scenario binaries and lets researchers author stress tests by moving or
inserting static obstacles and placing dynamic vehicles with custom poses, velocities,
trajectories, and goals (e.g., narrow corridors or parked cars); this isolates skills such as slalom maneuverability and
short-distance emergency braking. A \emph{real-time kinematics dashboard} plots the
realized metrics at every step against the OPM boundaries (Fig.~\ref{fig:editor-screenshot}),
highlighting comfort exceedances in red so it is immediately visible whether a breach
occurred in acceleration or jerk. Finally, \emph{multi-model auditing} compares two
rollouts either side by side, with synchronized timelines and feasible action boxes
$\feas(s)$, or overlaid on a single map, making visible where a policy under naive
clamping saturates its steering and leaves the lane while the adaptive model preserves
the authority to negotiate the obstacle.

\section{Results}

We now return to the two questions that motivated our experiments: whether enforcing the comfort envelope actually eliminates envelope-violating motion, and what it costs in navigation performance. We first report the aggregate numbers across all six kinematic models (Section~\ref{sec:results-quantitative}), then turn to the stress-test scenes that expose the mechanism behind those numbers (Section~\ref{sec:qualitative}).

\subsection{Quantitative Performance on PufferDrive}
\label{sec:results-quantitative}

\begin{figure*}[t]
  \centering
  \includegraphics[width=0.99\textwidth]{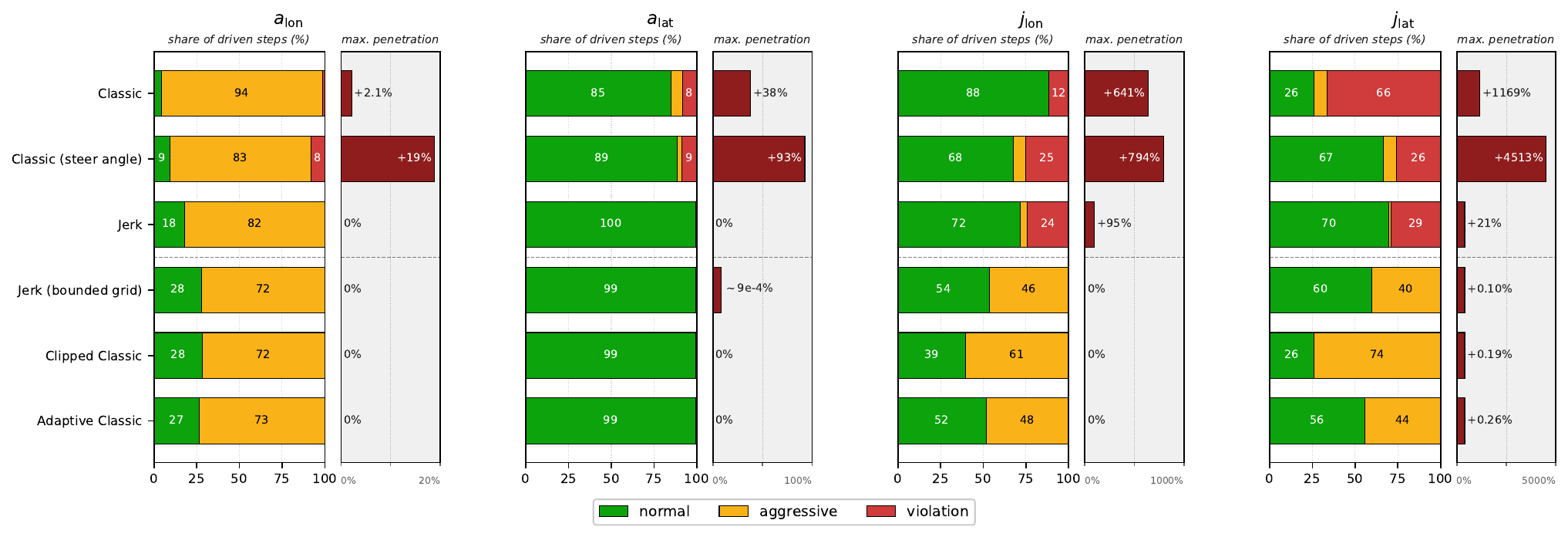}
  \caption{Driving-zone occupancy of the \emph{realized} kinematics: the share of driven steps whose realized quantity falls in the normal, aggressive, or violating band of the comfort envelope, per kinematic model (mean over three seeds). The constrained models enforce the \textsc{aggressive} comfort profile (bounded constraints); the unconstrained baselines enforce no limits. The adaptive model fills the aggressive band without ever leaving it; the unconstrained baselines place a large share of their mass outside the envelope. The shaded column adds how \emph{far} outside: the max penetration, i.e.\ the worst-case overshoot past the crossed bound relative to that bound ($+100\%$ = twice the limit). Its bars are proportional within a panel, each scaled to the axis printed below it.}
  \label{fig:zone_occupancy}
\end{figure*}

Table~\ref{tab:dynamics_models} reports navigation performance and the realized-kinematics violation rate for all six kinematic models over three seeds; Figure~\ref{fig:zone_occupancy} resolves the violation columns into the full driving-zone distribution.

\paragraph{Unconstrained control drives outside the envelope.}
The unconstrained \textsc{Classic} baselines achieve the highest goal completion performance, but only by driving outside the comfort limits. They exceed even the \textsc{aggressive} comfort envelope on far more than a quarter of their steps, mostly in lateral jerk, followed by longitudinal jerk and lateral acceleration, with maximum violations reaching 4513\% for jerk and 93\% for acceleration. In comparison, the constrained variants only exceed the bounds when e.g. the physical steering-angle limit prevents their enforcement, with maximum violations of just 0.26\% for jerk and $9\times10^{-4}$\% for acceleration. Which steering command the policy holds barely matters for the performance metrics, the steer-rate and steer-angle variants are within noise of each other in goal completion and collisions. Commanding the rate integrates the steering and thereby reduces the violations that a single abrupt angle change can produce. While the unconstrained \textsc{Jerk} baseline removes acceleration violations by construction, it worsens jerk violations because the action space of the jerk model is already partially outside the aggressive bounds.

\paragraph{Enforcing the envelope costs little.}
Both clipping and adaptive re-discretization reduce comfort violations to essentially zero (Fig.~\ref{fig:zone_occupancy}), while reducing goal completion by around two percentage points compared to the unconstrained Classic baseline. In contrast, the bounded-grid Jerk baseline achieves lower navigation performance, indicating that a single static action grid is overly restrictive across varying vehicle speeds. Among the per-step projection methods, adaptive re-discretization consistently performs best: it matches clipping's comfort guarantees while achieving slightly higher goal completion and lower collision and off-road rates under the aggressive profile. The same trend is maintained under the tighter normal profile.



\paragraph{The constraint does not make policies drive conservatively.}
Zero violations alone would be insufficient evidence if the policy simply avoided operating near the limits. Figure~\ref{fig:zone_occupancy} shows the opposite: the adaptive model spends $\approx 74\%$ of driven steps in the \emph{aggressive} longitudinal-acceleration band and roughly half in the aggressive jerk bands. It operates at the very edge of the safety envelope without crossing it; in contrast, the unconstrained baselines mostly fluctuate between normal operation and violation states for longitudinal and lateral jerk. 

\begin{table*}[t]
  \centering
  \caption{%
    Navigation performance per dynamics model on the PufferDrive validation split,
    under the \textsc{AGGRESSIVE} and \textsc{NORMAL} comfort profiles, averaged over three seeds, in percent.
    Unconstrained baselines do not enforce comfort limits and are reported once.
    Navigation columns give mean $\pm$ std. \textbf{Bold} marks the best value per column
    \emph{within each outer group} (Unconstrained, Aggressive, or Normal): the unconstrained
    baselines buy their navigation numbers with envelope-violating control, so comparing them against
    constrained models would be unfair.}
  \label{tab:dynamics_models}
  \footnotesize
  \setlength{\tabcolsep}{6pt}
  \begin{tabular}{@{}l ccc@{}}
    \toprule
    Dynamics model
      & Goal $\uparrow$ & Collisions \ $\downarrow$ & Offroad $\downarrow$ \\
    \midrule
    \multicolumn{4}{@{}l}{\emph{Unconstrained}} \\
    \textsc{Classic} & $98.74 \pm 0.03$ & $\mathbf{0.48 \pm 0.02}$ & $\mathbf{0.25 \pm 0.02}$ \\
    \textsc{Classic} (steer angle) & $\mathbf{98.79 \pm 0.03}$ & $0.49 \pm 0.02$ & $0.27 \pm 0.02$ \\
    \textsc{Jerk} & $97.68 \pm 0.02$ & $1.16 \pm 0.03$ & $0.53 \pm 0.04$ \\
    \midrule
    \multicolumn{4}{@{}l}{\emph{Aggressive}} \\
    \textsc{Jerk} (bounded grid) & $96.39 \pm 0.09$ & $1.79 \pm 0.04$ & $1.00 \pm 0.07$ \\
    Clipped \textsc{Classic} & $96.77 \pm 0.06$ & $1.68 \pm 0.02$ & $0.80 \pm 0.05$ \\
    Adaptive \textsc{Classic} & $\mathbf{96.89 \pm 0.04}$ & $\mathbf{1.65 \pm 0.03}$ & $\mathbf{0.71 \pm 0.02}$ \\
    \midrule
    \multicolumn{4}{@{}l}{\emph{Normal}} \\
    \textsc{Jerk} (bounded grid) & $92.32 \pm 0.08$ & $2.39 \pm 0.06$ & $1.27 \pm 0.07$ \\
    Clipped \textsc{Classic} & $92.51 \pm 0.03$ & $\mathbf{2.12 \pm 0.02}$ & $1.22 \pm 0.02$ \\
    Adaptive \textsc{Classic} & $\mathbf{92.73 \pm 0.04}$ & $2.27 \pm 0.01$ & $\mathbf{1.15 \pm 0.01}$ \\
    \bottomrule
  \end{tabular}
\end{table*}

\paragraph{Tightening the envelope.}
\label{sec:profile-trend}
Table~\ref{tab:dynamics_models} repeats the comparison under the tighter \textsc{normal} comfort profile, where the models may only command actions within the comfortable band. All families degrade by a similar margin ($\approx 4$ points of goal reaching) as the feasible set shrinks, and adaptive re-discretization remains the strongest constrained driver: the adaptive steer-rate model reaches $92.73\%$ goal completion, the best of any constrained model under the normal profile, while keeping violations at zero because it preserves full action resolution inside whatever box remains rather than collapsing a fixed grid onto its boundary. 

\subsection{Qualitative Analysis in Stress-Test Scenes}
\label{sec:qualitative}
Aggregate metrics conceal what happens where the limits bind. Inspecting rollouts in the hand-authored stress-test scenes (Section~\ref{sec:qualitative-setup}) with \toolname{}'s dashboard (Fig.~\ref{fig:editor-screenshot}), the unconstrained configurations show acceleration and jerk spikes leaving the envelope just before sharp turns or collisions, while our adaptive model keeps inputs smooth and strictly inside the bounds. Figure~\ref{fig:jerk_failure} makes this concrete on a slalom under four models: the bounded-grid \textsc{Jerk} baseline deviates from the reference and leaves its goals unreached, whereas our adaptive model tracks the route through the slalom and reaches its goals cleanly within the comfort bounds.

\newcommand{\jerkfailpanel}[1]{%
  \includegraphics[angle=0, trim=370bp 380bp 160bp 140bp, clip, width=\textwidth]{#1}}
\begin{figure*}[t]
\centering
\begin{subfigure}[b]{0.23\textwidth}
  \centering
  \jerkfailpanel{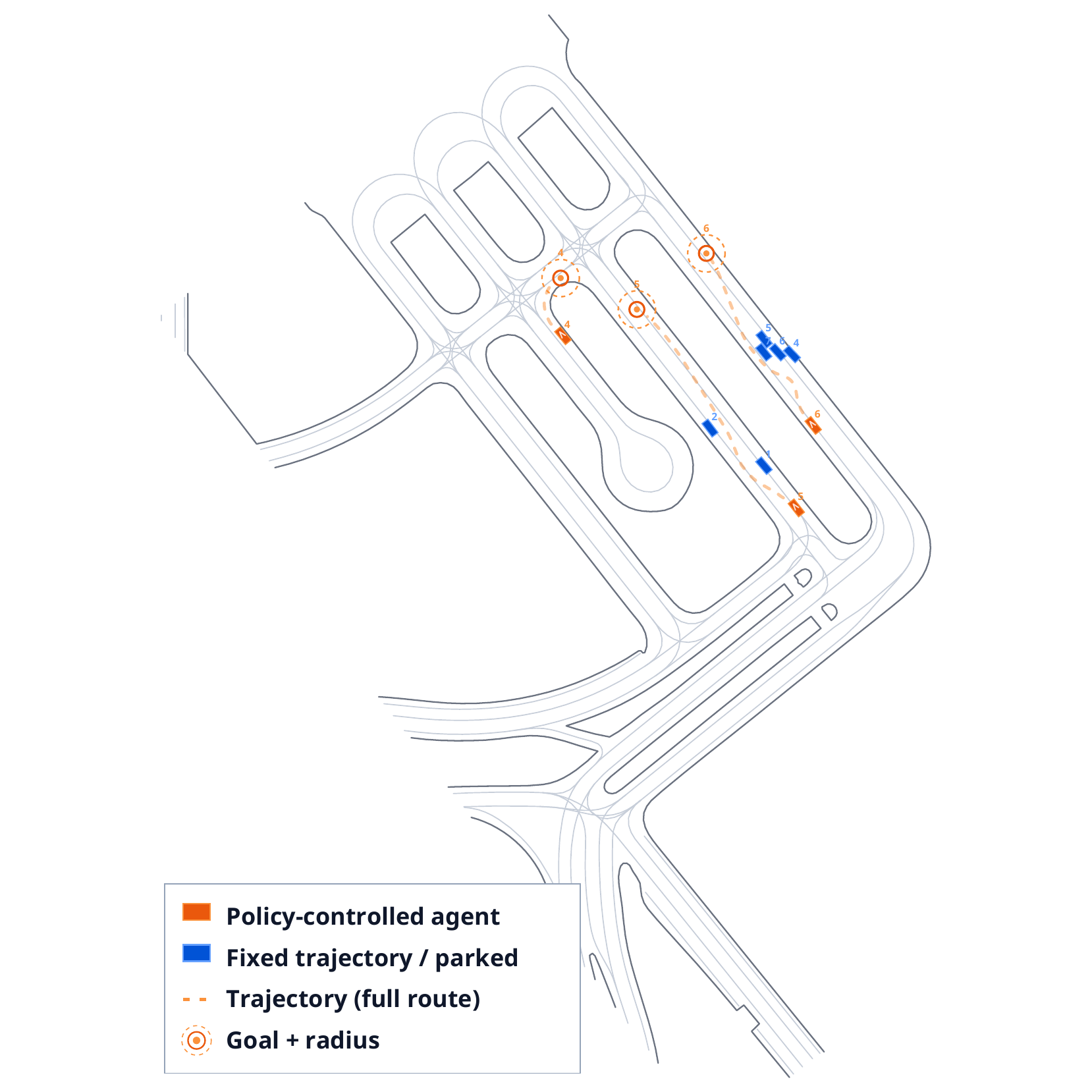}
  \caption{\textsc{Classic} (steer angle).}
  \label{fig:jerk_failure_classic}
\end{subfigure}%
\hfill
\begin{subfigure}[b]{0.23\textwidth}
  \centering
  \jerkfailpanel{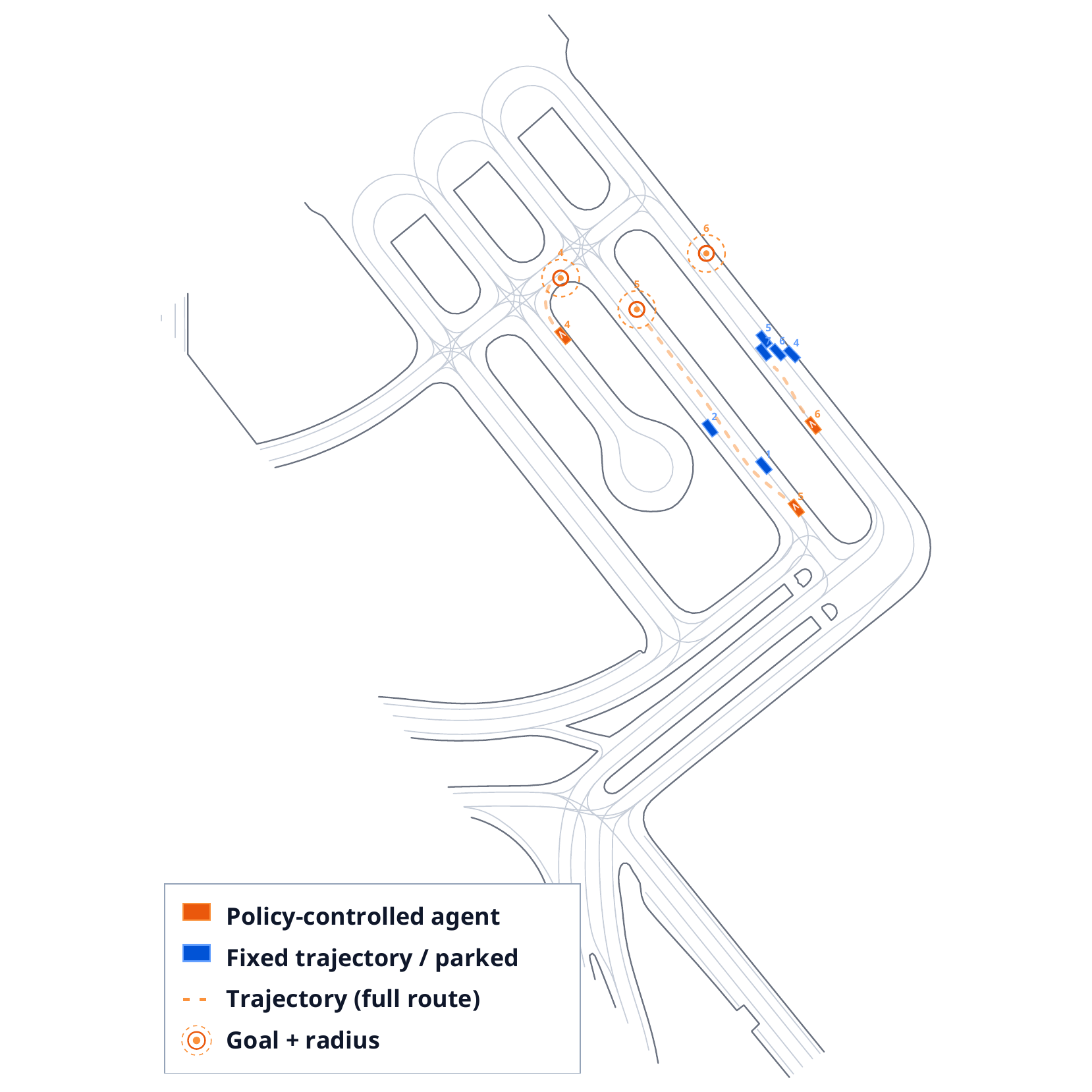}
  \caption{Clipped.}
  \label{fig:jerk_failure_clipped}
\end{subfigure}%
\hfill
\begin{subfigure}[b]{0.23\textwidth}
  \centering
  \jerkfailpanel{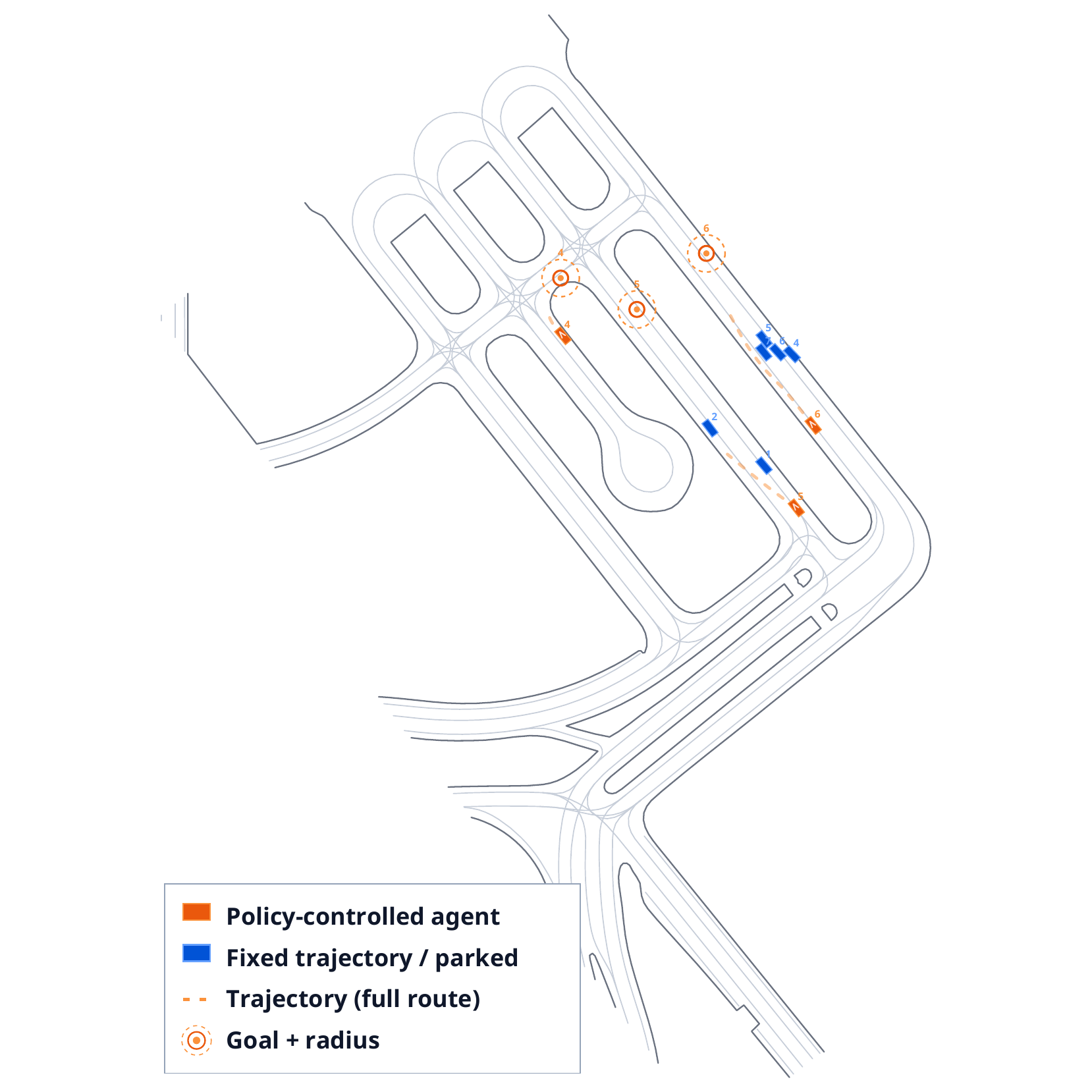}
  \caption{\textsc{bounded Jerk}.}
  \label{fig:jerk_failure_jerk}
\end{subfigure}%
\hfill
\begin{subfigure}[b]{0.23\textwidth}
  \centering
  \jerkfailpanel{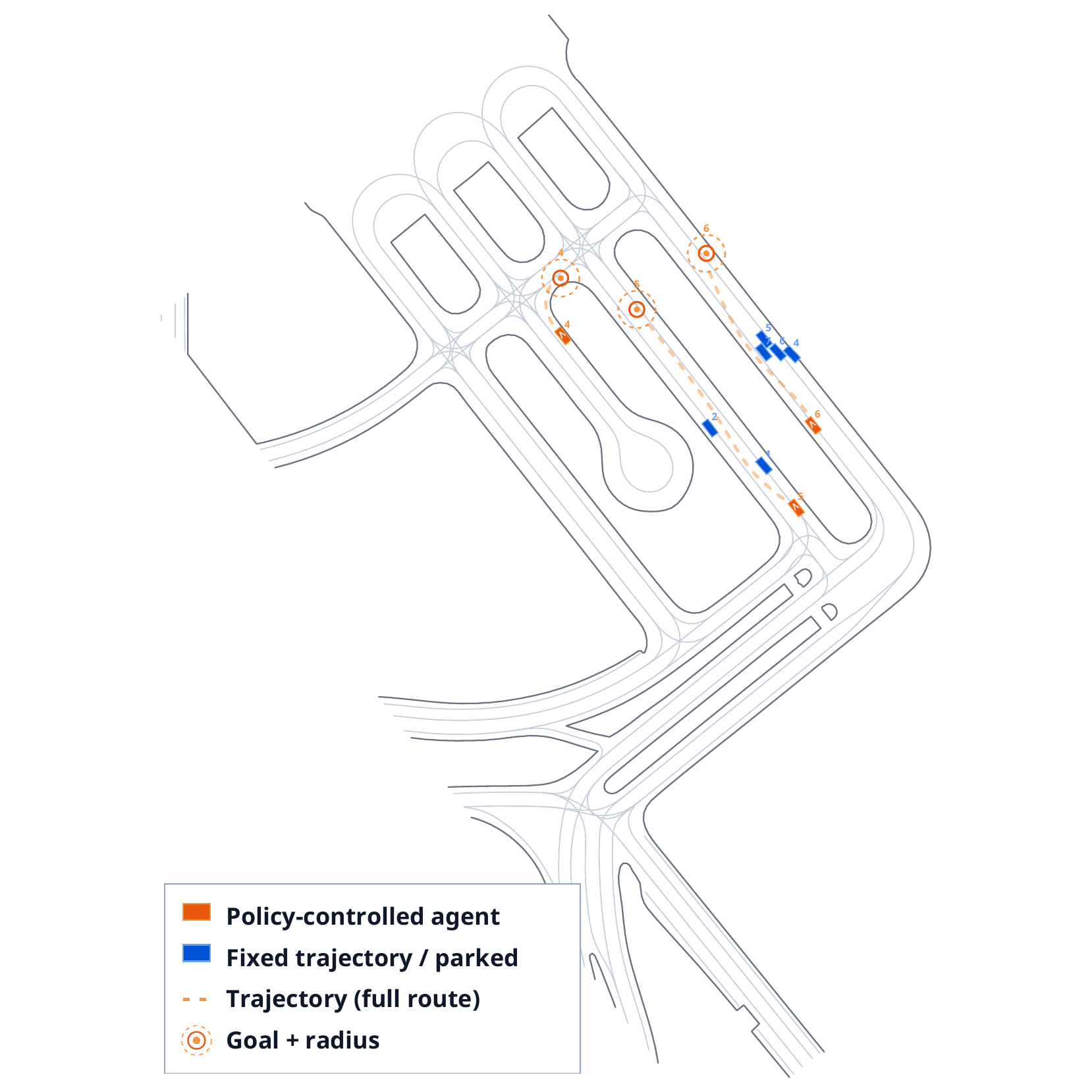}
  \caption{Adaptive (ours).}
  \label{fig:jerk_failure_adaptive}
\end{subfigure}
\caption{The same stress-test scenes under four kinematic models, zoomed to the identical region around the policy-controlled route. Policy-controlled agents are orange (numbered 4--6, dashed routes; goals shown as numbered ringed dots); parked agents are blue. Our adaptive model (d) tracks the reference through the scene and reaches all goals within the comfort bounds, whereas the bounded-grid \textsc{Jerk} baseline (c) deviates from the route and leaves its goals unreached (empty rings); the unconstrained \textsc{Classic} (steer angle) (a) and clipped (b) variants are shown for reference.}
\label{fig:jerk_failure}
\end{figure*}

\section{Discussion}
\label{sec:discussion}

The quantitative results and qualitative profiles validate our hypotheses regarding RL action spaces under constraints, yielding several insights.

\paragraph{Comfort vs.\ navigability trade-off.}
Moving from the \textsc{aggressive} to the \textsc{normal} profile consistently drops goal completion by roughly $4$ percentage points and raises collisions, as the tighter bounds  curb aggressive braking and acceleration: passenger comfort directly constrains defensive reactivity. One might worry that bounding the action space removes the emergency maneuvers a policy needs to stay safe, but the envelope is a configurable profile rather than a fixed ceiling. The \textsc{aggressive} band sits at the extremes of human driving, so the policy keeps the full range a human would use and forgoes only motion beyond it, and a deployment wanting a wider margin can enforce a looser profile. Empirically the constraint rarely binds, costing under two points of goal completion against the unconstrained baseline while removing nearly all violations.

\paragraph{Adaptive mapping beats clipping and jerk control.}
Adaptive mapping outperforms clipped steering and jerk control, mainly in safety-critical navigation. While all constrained methods achieve similar goal completion, adaptive re-discretization preserves steering resolution within the feasible range, improving control authority when avoiding collisions and off-road events. Clipping instead saturates the fixed action grid as the steering range shrinks at high speed. Jerk control is weakest in these hard-navigation settings because second-order actions introduce an additional integration chain ($j \rightarrow a \rightarrow v \rightarrow x$), making the control problem harder to learn and credit assignment more difficult. Thus, while jerk control is simple to constrain, direct first-order control is better suited for precise obstacle avoidance and boundary recovery.

\paragraph{Action-space design: rate vs.\ angle, hard vs.\ soft.}
We command steering rate ($\dot\delta$) rather than steering angle ($\delta$) by default, and all constrained \textsc{Classic} variants use it. Rate control provides a useful balance between responsiveness and smoothness: it avoids abrupt steering changes while retaining more control authority than jerk-based control, which introduces an additional integration chain. The two unconstrained variants are compared in Table~\ref{tab:dynamics_models} to isolate the effect of the steering action space where both perform comparably bad. Alternatively, comfort can be encouraged through reward penalties, as Gigaflow~\cite{cusumano2025selfplay} does for harsh acceleration and jerk. Such a penalty prices comfort against other objectives through a tunable weight rather than ruling a maneuver out, so the optimal policy still accepts a violation whenever the payoff of an evasive maneuver exceeds its cost. Reward design therefore shapes driving preferences but cannot reliably define which actions should be considered unacceptable. Hard action constraints and reward shaping are complementary: the constraint makes out-of-envelope motion unreachable by construction and so guarantees compliance with the modeled comfort envelope~\cite{alshiekh2018shielding,dalal2018safe}, while the reward can optimize driving style within the feasible region.

\subsection{Conclusion}
\label{sec:discussion-conclusion}
We introduced a constraint-aware, adaptive action-mapping framework to address the reliance of learned driving policies on maneuvers no human driver would perform and no occupant would accept. By analytically inverting the lateral-jerk constraint in next-steering-angle space, our approach dynamically scales the control grid to span exactly the comfort envelope at each step. On WOMD and a hand-authored slalom stress test, the proposed adaptive steer-rate \textsc{Classic} model enforced comfort compliance (limiting comfort violations to below $0.01\%$) and outperformed clipped and direct-jerk baselines in navigability. Because every action the policy can select now lies inside the enforced comfort envelope, the resulting safety metrics reflect policy quality rather than the simulator's permissiveness, closing the gap that motivated this work. Interactive audits via \toolname{} confirmed that adaptive re-discretization preserves steering authority and prevents grid collapse, allowing the policy to negotiate the slalom's tight turns and narrow gaps.


\section*{Acknowledgement}
This work is part of BrainLinks-BrainTools which is funded by the Federal Ministry of Economics, Science and Arts of Baden-Württemberg within the sustainability program for projects of the excellence initiative II.
Also, this research was funded by the Deutsche Forschungsgemeinschaft (DFG, German Research Foundation) under grant number 539134284, through EFRE (FEIH\_2698644) and the state of Baden-Württemberg. 

\begin{center}
\raisebox{-0.5\height}{\includegraphics[width=0.42\columnwidth]{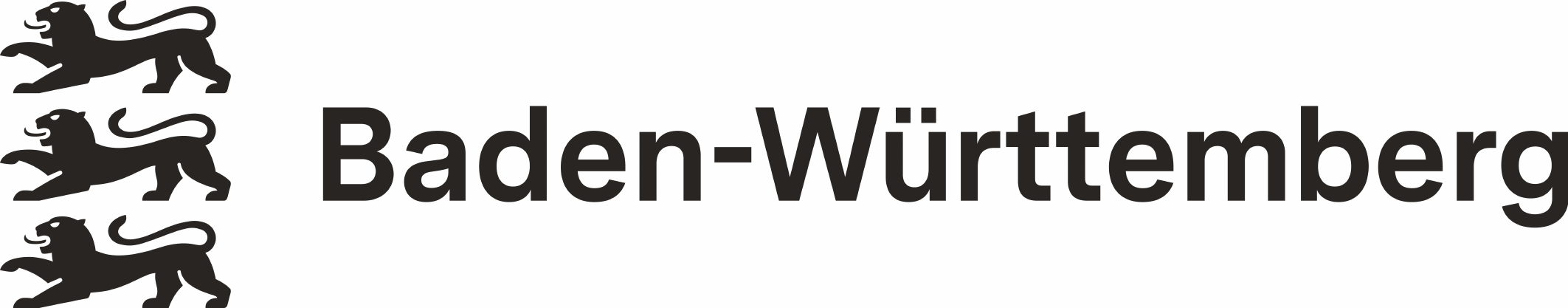}}
\hspace{0.04\columnwidth}
\raisebox{-0.5\height}{\includegraphics[width=0.42\columnwidth]{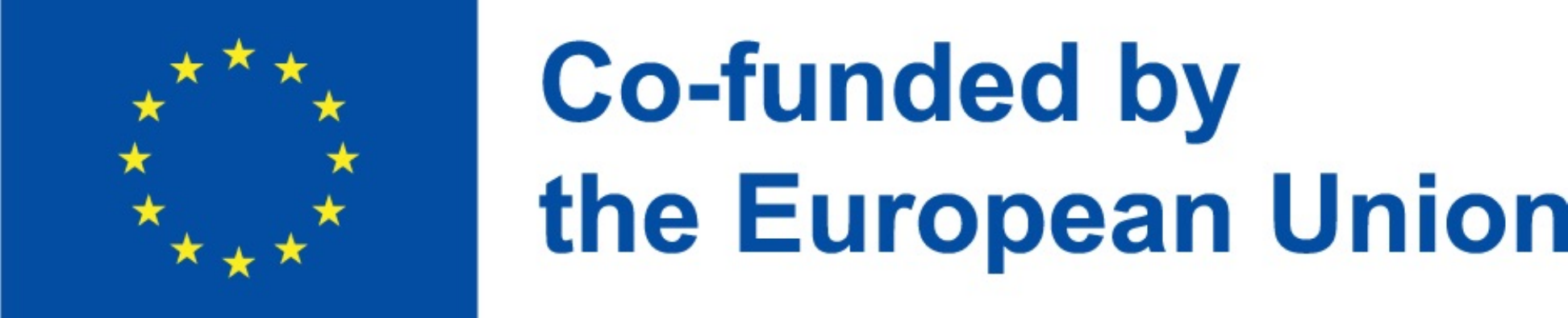}}
\end{center}

%
%
\bibliographystyle{splncs04}
\bibliography{main}

@String(ICCV  = {Int. Conf. Comput. Vis.})

@String(ECCV  = {Eur. Conf. Comput. Vis.})

@String(NeurIPS = {Adv. Neural Inform. Process. Syst.})

@String(ICML  = {Int. Conf. Mach. Learn.})

@String(ICLR  = {Int. Conf. Learn. Represent.})

@String(AAAI  = {AAAI})

@String(ICCV  = {ICCV})

@String(ECCV  = {ECCV})

@String(NeurIPS = {NeurIPS})

@String(ICML  = {ICML})

@String(ICLR  = {ICLR})

@inproceedings{ettinger2021womd,
  author    = {Ettinger, Scott and Cheng, Shuyang and Caine, Benjamin and Liu, Chenxi and Zhao, Hang and Pradhan, Sabeek and Chai, Yuning and Sapp, Ben and Qi, Charles R. and Zhou, Yin and Yang, Zoey and Chouard, Aur{\'e}lien and Sun, Pei and Ngiam, Jiquan and Vasudevan, Vijay and McCauley, Alexander and Shlens, Jonathon and Anguelov, Dragomir},
  title     = {Large Scale Interactive Motion Forecasting for Autonomous Driving: The Waymo Open Motion Dataset},
  booktitle = ICCV,
  pages     = {9690--9699},
  year      = {2021}
}

@inproceedings{vinitsky2022nocturne,
  author    = {Vinitsky, Eugene and Lichtl{\'e}, Nathan and Yang, Xiaomeng and Amos, Brandon and Foerster, Jakob},
  title     = {Nocturne: A Scalable Driving Benchmark for Bringing Multi-Agent Learning One Step Closer to the Real World},
  booktitle = NeurIPS,
  year      = {2022}
}

@inproceedings{gulino2023waymax,
  author    = {Gulino, Cole and Fu, Justin and Luo, Wenjie and Tucker, George and Bronstein, Eli and Lu, Yiren and Harb, Jean and Pan, Xinlei and Wang, Yan and Chen, Xiangyu and Co-Reyes, John D. and Agarwal, Rishabh and Roelofs, Rebecca and Lu, Yao and Montali, Nico and Mougin, Paul and Yang, Zoey and White, Brandyn and Faust, Aleksandra and McAllister, Rowan and Anguelov, Dragomir and Sapp, Benjamin},
  title     = {Waymax: An Accelerated, Data-Driven Simulator for Large-Scale Autonomous Driving Research},
  booktitle = NeurIPS,
  year      = {2023}
}

@inproceedings{kazemkhani2024gpudrive,
  author    = {Kazemkhani, Saman and Pandya, Aarav and Cornelisse, Daphne and Shacklett, Brennan and Vinitsky, Eugene},
  title     = {{GPUDrive}: Data-Driven, Multi-Agent Driving Simulation at 1 Million {FPS}},
  booktitle = ICLR,
  year      = {2025}
}

@inproceedings{cusumano2025selfplay,
  author    = {Cusumano-Towner, Marco and Hafner, David and Hertzberg, Alex and Huval, Brody and Petrenko, Aleksei and Vinitsky, Eugene and Wijmans, Erik and Killian, Taylor and Bowers, Stuart and Sener, Ozan and Koltun, Vladlen and Kr{\"a}henb{\"u}hl, Philipp},
  title     = {Robust Autonomy Emerges from Self-Play},
  booktitle = ICML,
  series    = {Proceedings of Machine Learning Research (PMLR)},
  volume    = {267},
  pages     = {11710--11737},
  year      = {2025}
}

@book{rajamani2011vehicle,
  author    = {Rajamani, Rajesh},
  title     = {Vehicle Dynamics and Control},
  edition   = {2nd},
  publisher = {Springer},
  series    = {Mechanical Engineering Series},
  year      = {2012},
  doi       = {10.1007/978-1-4614-1433-9}
}

@inproceedings{kong2015kinematic,
  author    = {Kong, Jason and Pfeiffer, Mark and Schildbach, Georg and Borrelli, Francesco},
  title     = {Kinematic and Dynamic Vehicle Models for Autonomous Driving Control Design},
  booktitle = {IEEE Intelligent Vehicles Symposium (IV)},
  pages     = {1094--1099},
  year      = {2015}
}

@inproceedings{polack2017kinematic,
  author    = {Polack, Philip and Altch{\'e}, Florent and d'Andr{\'e}a-Novel, Brigitte and de La Fortelle, Arnaud},
  title     = {The Kinematic Bicycle Model: A Consistent Model for Planning Feasible Trajectories for Autonomous Vehicles?},
  booktitle = {IEEE Intelligent Vehicles Symposium (IV)},
  pages     = {812--818},
  year      = {2017}
}

@inproceedings{bae2020comfort,
  author    = {Bae, Il and Moon, Jaeyoung and Jhung, Junekyo and Suk, Ho and Kim, Taewoo and Park, Hyungbin and Cha, Jaekwang and Kim, Jinhyuk and Kim, Dohyun and Kim, Shiho},
  title     = {Self-Driving like a Human Driver instead of a Robocar: Personalized Comfortable Driving Experience for Autonomous Vehicles},
  booktitle = {Machine Learning for Autonomous Driving Workshop at the 33rd Conference on Neural Information Processing Systems (NeurIPS)},
  address   = {Vancouver, Canada},
  year      = {2019}
}

@article{feng2017jerk,
  author  = {Feng, Fred and Bao, Shan and Sayer, James R. and Flannagan, Carol and Manser, Michael and Wunderlich, Robert},
  title   = {Can vehicle longitudinal jerk be used to identify aggressive drivers? An examination using naturalistic driving data},
  journal = {Accident Analysis \& Prevention},
  volume  = {104},
  pages   = {125--136},
  year    = {2017},
  doi     = {10.1016/j.aap.2017.04.012}
}

@inproceedings{suarez2024pufferlib,
  author    = {Su{\'a}rez, Joseph},
  title     = {{PufferLib}: Making Reinforcement Learning Libraries and Environments Play Nice},
  booktitle = {Agent Learning in Open-Endedness Workshop at the Conference on Neural Information Processing Systems (NeurIPS)},
  year      = {2023}
}

@inproceedings{montali2023wosac,
  author    = {Montali, Nico and Lambert, John and Mougin, Paul and Kuefler, Alex and Rhinehart, Nicholas and Li, Michelle and Gulino, Cole and Emrich, Tristan and Yang, Zoey and Whiteson, Shimon and White, Brandyn and Anguelov, Dragomir},
  title     = {The Waymo Open Sim Agents Challenge},
  booktitle = NeurIPS,
  year      = {2023}
}

@inproceedings{cui2020deepkinematic,
  author    = {Cui, Henggang and Nguyen, Thi and Chou, Fang-Chieh and Lin, Tsung-Han and Schneider, Jeff and Bradley, David and Djuric, Nemanja},
  title     = {Deep Kinematic Models for Kinematically Feasible Vehicle Trajectory Predictions},
  booktitle = {IEEE Int. Conf. Robot. Autom. (ICRA)},
  pages     = {10563--10569},
  year      = {2020}
}

@inproceedings{salzmann2020trajectronpp,
  author    = {Salzmann, Tim and Ivanovic, Boris and Chakravarty, Punarjay and Pavone, Marco},
  title     = {Trajectron++: Dynamically-Feasible Trajectory Forecasting with Heterogeneous Data},
  booktitle = ECCV,
  pages     = {683--700},
  year      = {2020}
}

@article{westny2024stochastic,
  author  = {Zheng, Laura and Son, Sanghyun and Liang, Jing and Wang, Xijun and Clipp, Brian and Lin, Ming C.},
  title   = {Deep Stochastic Kinematic Models for Probabilistic Motion Forecasting in Traffic},
  journal = {arXiv preprint arXiv:2406.01431},
  year    = {2024}
}

@inproceedings{krasowski2020safe,
  author    = {Krasowski, Hanna and Wang, Xiao and Althoff, Matthias},
  title     = {Safe Reinforcement Learning for Autonomous Lane Changing Using Set-Based Prediction},
  booktitle = {IEEE Int. Conf. Intell. Transp. Syst. (ITSC)},
  year      = {2020}
}

@inproceedings{wang2021commonroadrl,
  author    = {Wang, Xiao and Krasowski, Hanna and Althoff, Matthias},
  title     = {{CommonRoad-RL}: A Configurable Reinforcement Learning Environment for Motion Planning of Autonomous Vehicles},
  booktitle = {IEEE Int. Conf. Intell. Transp. Syst. (ITSC)},
  year      = {2021}
}

@article{dalal2018safe,
  author  = {Dalal, Gal and Dvijotham, Krishnamurthy and Vecerik, Matej and Hester, Todd and Paduraru, Cosmin and Tassa, Yuval},
  title   = {Safe Exploration in Continuous Action Spaces},
  journal = {arXiv preprint arXiv:1801.08757},
  year    = {2018}
}

@inproceedings{alshiekh2018shielding,
  author    = {Alshiekh, Mohammed and Bloem, Roderick and Ehlers, R{\"u}diger and K{\"o}nighofer, Bettina and Niekum, Scott and Topcu, Ufuk},
  title     = {Safe Reinforcement Learning via Shielding},
  booktitle = AAAI,
  year      = {2018}
}

@inproceedings{cheng2019cbf,
  author    = {Cheng, Richard and Orosz, G{\'a}bor and Murray, Richard M. and Burdick, Joel W.},
  title     = {End-to-End Safe Reinforcement Learning through Barrier Functions for Safety-Critical Continuous Control Tasks},
  booktitle = AAAI,
  pages     = {3387--3395},
  year      = {2019}
}

@article{theile2025actionmapping,
  author  = {Theile, Mirco and Dirnberger, Lukas and Trumpp, Raphael and Caccamo, Marco and Sangiovanni-Vincentelli, Alberto L.},
  title   = {Action Mapping for Reinforcement Learning in Continuous Environments with Constraints},
  journal = {Reinforcement Learning Journal},
  volume  = {6},
  year    = {2025}
}

@article{lunardi2024jerkbounded,
  author  = {Alizadeh Kolagar, Seyed Adel and Heydari Shahna, Mehdi and Mattila, Jouni},
  title   = {Combining Deep Reinforcement Learning with a Jerk-Bounded Trajectory Generator for Kinematically Constrained Motion Planning},
  journal = {arXiv preprint arXiv:2410.20907},
  year    = {2024}
}

@article{dewinkel2023comfort,
  author  = {de Winkel, Ksander N. and Irmak, Tugrul and Happee, Riender and Shyrokau, Barys},
  title   = {Standards for Passenger Comfort in Automated Vehicles: Acceleration and Jerk},
  journal = {Applied Ergonomics},
  volume  = {106},
  pages   = {103881},
  year    = {2023}
}

@inproceedings{xu2023bits,
  author    = {Xu, Danfei and Chen, Yuxiao and Ivanovic, Boris and Pavone, Marco},
  title     = {{BITS}: Bi-level Imitation for Traffic Simulation},
  booktitle = {IEEE Int. Conf. Robot. Autom. (ICRA)},
  pages     = {2929--2936},
  year      = {2023}
}

@article{elbanhawi2015comfort,
  author  = {Elbanhawi, Mohamed and Simic, Milan and Jazar, Reza},
  title   = {In the Passenger Seat: Investigating Ride Comfort Measures in Autonomous Cars},
  journal = {IEEE Intelligent Transportation Systems Magazine},
  volume  = {7},
  number  = {3},
  pages   = {4--17},
  year    = {2015}
}

@inproceedings{martin2008jerk,
  author    = {Martin, Daniel and Litwhiler, Dale},
  title     = {An Investigation of Acceleration and Jerk Profiles of Public Transportation Vehicles},
  booktitle = {ASEE Annual Conference \& Exposition},
  year      = {2008}
}

@article{schulman2017proximal,
  title     = {Proximal policy optimization algorithms},
  author    = {Schulman, John and Wolski, Filip and Dhariwal, Prafulla and Radford, Alec and Klimov, Oleg},
  journal   = {arXiv preprint arXiv:1707.06347},
  year      = {2017}
}

%

\end{document}